\documentclass[letterpaper]{article} 
\usepackage{aaai2026}  
\usepackage{times}  
\usepackage{helvet}  
\usepackage{courier}  
\usepackage[hyphens]{url}  
\usepackage{graphicx} 
\usepackage{natbib}  
\usepackage{caption} 
\usepackage{algorithm}
\usepackage{algorithmic}

\usepackage{newfloat}
\usepackage{listings}
\DeclareCaptionStyle{ruled}{labelfont=normalfont,labelsep=colon,strut=off} 
\floatstyle{ruled}
\newfloat{listing}{tb}{lst}{}
\floatname{listing}{Listing}
\usepackage[utf8]{inputenc} 
\usepackage{booktabs}       
\usepackage{amsfonts}       
\usepackage{nicefrac}       
\usepackage{microtype}      
\usepackage{xcolor}         
\usepackage{multirow}    
\usepackage{tabularx}    
\usepackage{siunitx}   
\usepackage{amsmath}
\usepackage{subcaption}
\usepackage{arydshln}  
\usepackage{colortbl}
\usepackage{makecell}

\title{Wi-CBR: Salient-aware Adaptive WiFi Sensing for Cross-domain \\ Behavior Recognition}
\author{
    Ruobei Zhang$^1$, Shengeng Tang$^1$\thanks{Corresponding authors.}, Huan Yan$^2$, Xiang Zhang$^3$, Jiabao Guo$^{1*}$
}
\affiliations{
    $^1$ Hefei University of Technology\\
    $^2$ Guizhou Normal University\\
    $^3$ Tianjin University\\[2pt]

    zrb@mail.hfut.edu.cn, \{tangsg, garbo\_guo\}@hfut.edu.cn, yh1995.cs@gmail.com, zhangxiang@ieee.org
}

\usepackage{bibentry}

\begin{document}

\maketitle

\begin{abstract}
The challenge in WiFi-based cross-domain Behavior Recognition lies in the significant interference of domain-specific signals on gesture variation. However, previous methods alleviate this interference by mapping the phase from multiple domains into a common feature space. If the Doppler Frequency Shift (DFS) signal is used to dynamically supplement the phase features to achieve better generalization, it enables the model to not only explore a wider feature space but also to avoid potential degradation of gesture semantic information. Specifically, we propose a novel \textit{Salient-aware Adaptive \textbf{Wi}Fi Sensing for \textbf{C}ross-domain \textbf{B}ehavior \textbf{R}ecognition}  (\textbf{Wi-CBR}), which constructs a dual-branch self-attention module that captures temporal features from phase information reflecting dynamic path length variations while extracting kinematic features from DFS correlated with motion velocity. Moreover, we design a Saliency Guidance Module that employs group attention mechanisms to mine critical activity features and utilizes gating mechanisms to optimize information entropy, facilitating feature fusion and enabling effective interaction between salient and non-salient behavioral characteristics. Extensive experiments on two large-scale public datasets (Widar3.0 and XRF55) demonstrate the superior performance of our method in both in-domain and cross-domain scenarios.
\end{abstract}

\begin{links}
    \link{Code}{https://github.com/zrbwsw/Wi-CBR}
\end{links}

\section{Introduction}

Human Behavior Recognition (HBR) enables systems to intelligently interpret human actions and is widely applied in areas like human-computer interaction~\cite{tang2025discrete, tang2025sign}, Interactive integration~\cite{song2024emotional, song2023emotion}, action understanding~\cite{xu2025towards, zhang2025temporal}, and accessible communication~\cite{tangslt2022, tang2022gloss}. WiFi-based HBR, leveraging Channel State Information (CSI) and Received Signal Strength Indicator (RSSI), has gained attention for its ability to model motion-induced signal changes~\cite{1data2,1data3}. While coarse-grained activities are easily detectable, fine-grained gestures remain challenging due to subtle movements~\cite{14}. By mapping CSI variations to behaviors, WiFi sensing supports applications in homes, workplaces, and other indoor environments~\cite{Wigest,WiFinger}.

\begin{figure}[t] 
  \centering
  \includegraphics[width=1\linewidth]{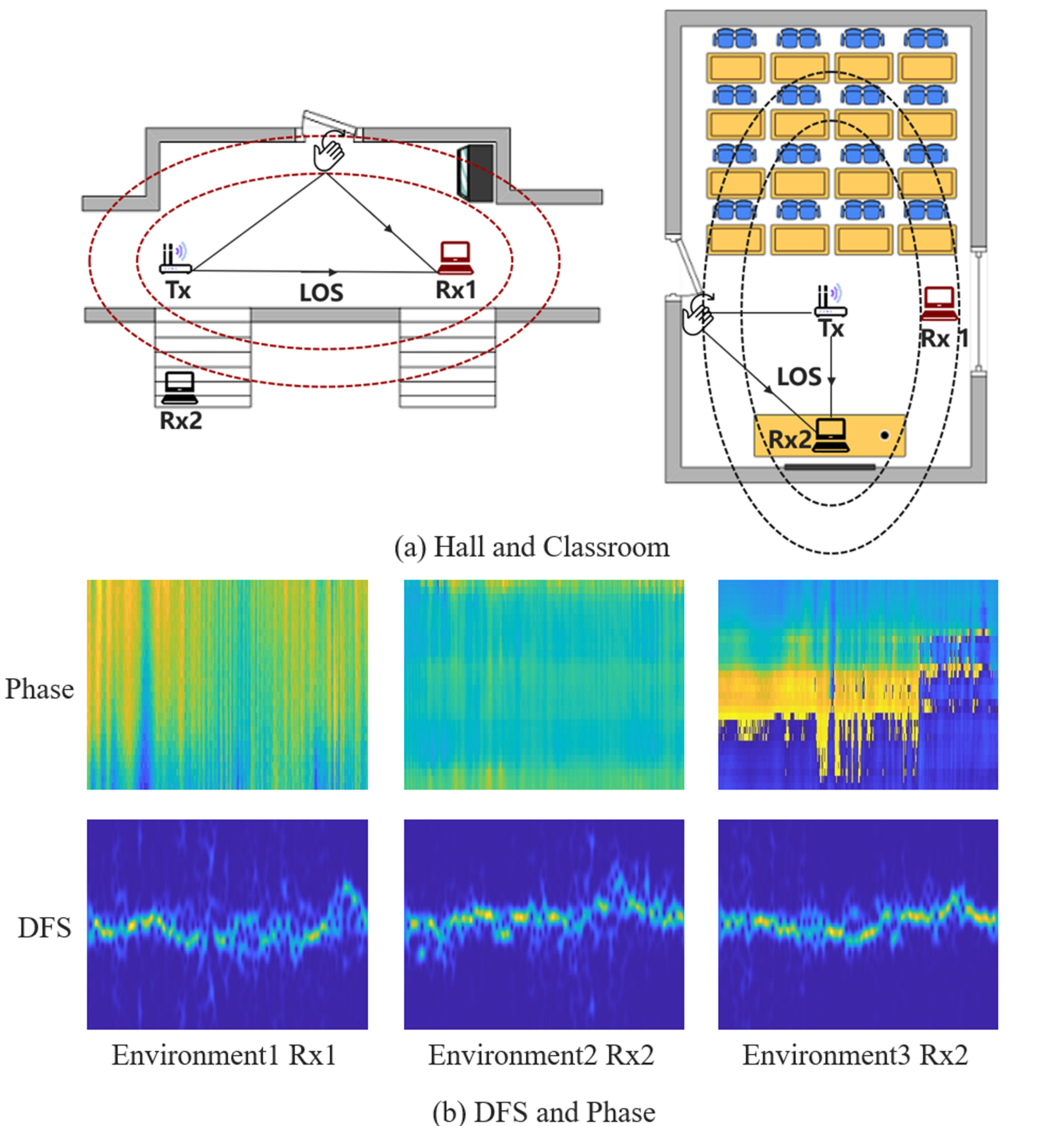} 
  \caption{When the user and behavior remain unchanged, relative to the location of transmitter–receiver (Tx-Rx) pairs or the overall environment changes, the phase changes significantly, and DFS has a certain degree of domain independence.}
  \label{fig:cross_domain}
\end{figure}

An important challenge in WiFi-based behavior recognition is environmental dependence, as wireless signals often carry environment-specific information that is unrelated to human behavior\cite{zhang2025wiopen}. Cross-domain recognition accuracy significantly decreases due to differences in data distribution, noise, and changes in the propagation path. Traditional methods are divided into Modeling-Based and Learning-Based approaches~\cite{2modelandlearn1,2modelandlearn2}. Modeling-Based methods use handcrafted features that are robust to environmental variation but lose raw signal richness, limiting performance in complex tasks. Learning-Based methods leverage deep neural networks to learn directly from raw CSI but are hindered by environmental sensitivity and a lack of domain-invariant guidance. Therefore, how to integrate original features with domain-independent features is key to achieving effective cross-domain generalization~\cite{guo2024style, zhang2025style, wu2025percept, yin2022mix, guo2025domain}, bridging the gap between controlled training environments and real-world applications.

Our observations find that, in wireless communication systems, the characteristics of the received signal are influenced by various factors, including the positions of the transmitter (Tx) and receiver (Rx), the transmission path, and the dynamics of the overall environment. As shown in Fig.~\ref{fig:cross_domain}, significant variations in phase values occur when the receiver's position changes, even if the user and behavior remain constant. Figure (b) further demonstrates that phase values are highly sensitive to environmental changes, exhibiting unstable patterns across different environments and receiver positions. In contrast, the DFS shows greater domain independence and stability, with its features generally remaining consistent across varying environments and receiver locations. Building on these observations, the proposed method leverages the stability of DFS features while addressing the sensitivity of phase values. By extracting and aligning domain-invariant DFS features, the approach enhances the adaptability to dynamic and multi-domain scenarios. Furthermore, we aim to preserve the rich temporal-spatial patterns of raw signals while embedding domain-independent priors that facilitate robust cross-domain generalization. This hybrid motivation directly informs the network design, enabling complementary feature learning and improving cross-domain performance.

In this work, we propose Wi-CBR to leverage the full spectrum of raw data while ensuring the integration of domain-independent features. Specifically, our method introduces a novel multimodal collaborative awareness framework that efficiently combines phase data, which captures dynamic path length changes, and DFS data, which reflects frequency shifts tied to gesture speed. To capture spatial-temporal patterns within each modality, we employ a Two-branch self-attention module, enabling the system to focus on important temporal and spatial features within each signal type. A group attention mechanism is then applied to the concatenated phase and DFS features, allowing the model to identify key group features that are essential for behavior recognition. Finally, a gating mechanism is used to divide the fused features into enhancement and suppression branches, optimizing information entropy and facilitating collaborative complementarity. This fusion of multiple data sources, along with the innovative use of attention and gating mechanisms, enables more accurate and robust behavior recognition, particularly in cross-domain scenarios. Our main contributions are summarized as follows:

\begin{itemize}
    \item We propose Wi-CBR, a novel Salient-aware Adaptive WiFi Sensing framework for Cross-domain Behavior Recognition. Wi-CBR proposes a two-branch self-attention module that captures temporal features from phase information, reflecting dynamic path length variations, while extracting kinematic features from DFS, which are correlated with motion velocity.
    
    \item We design a Saliency Guidance Module that leverages group attention mechanisms to identify critical activity features. This module incorporates gating mechanisms to optimize information entropy, facilitating feature fusion and enabling effective interaction between salient and non-salient behavioral characteristics.

    \item Extensive experiments on two large-scale public datasets demonstrate the superior performance of our method in both in-domain and cross-domain scenarios.
\end{itemize}

\section{Related Work}

\subsection{Modeling-Based HBR}

Modeling-based approaches preprocess raw CSI data, extract manual features (e.g., velocity statistics), and use machine learning for gesture recognition~\cite{WiFinger,WIHGR,2model6,2model3,2model2,Wi-NN}. WiGest~\cite{Wigest} relies on coarse-grained RSS, limiting accuracy. WiMU~\cite{WiMU} struggles with scalability, and WiDraw~\cite{Widraw} requires over 25 transceivers, making it impractical. QGesture~\cite{Qgesture} uses two antennas but depends on prior hand position knowledge. \cite{2model3} introduces dynamic phase exponential error for gesture quality, while Wi-NN\cite{Wi-NN} applies time-domain feature selection with KNN classification. These methods link WiFi signals to gestures but neglect environmental impact, as gesture performance in varying environments alters WiFi waveforms~\cite{Chen2023, Kang2021}. Widar 3.0~\cite{Widar3.0} introduces domain-agnostic BVP features, and WiHF~\cite{WIHF} derives domain-independent motion patterns; however, handcrafted features limit the capture of spatiotemporal cues. WiGesture~\cite{WiGesture} focuses on position-independent Motion Navigation Primitives (MNP). WiGNN\cite{WiGNN} focuses on graph modeling for multi-receiver topologies through GNN-based temporal-frequency aggregation and data augmentation.

\subsection{Learning-Based HBR}

Learning-based methods directly process raw CSI data, such as amplitude and phase, for automatic pattern recognition. Wikey~\cite{Wikey} enables keystroke recognition but is highly sensitive to environmental changes. WiSign~\cite{WiSign} extracts spatio-temporal features for sign language recognition but requires extensive domain-specific training. Tong et al.\cite{Tong2023} introduced a CNN-GRU-Attention (CGA) model with phase correction and gesture truncation for improved data validity, while Yang et al.\cite{Yang2019} proposed a CNN-RNN architecture for enhanced spatiotemporal pattern learning. WiHGR~\cite{WIHGR} uses a phase difference matrix and an improved ABGRU for feature extraction. To address cross-domain challenges, CROSSGR~\cite{CrossGR} extracts gesture-related features independent of users. WiGr~\cite{WiGr} uses query-class prototype similarity to mitigate cross-domain variations. WIGRUNT~\cite{WIGRUNT} applies a spatio-temporal dual-attention network with ResNet for feature extraction, while Wi-SFDAGR~\cite{Wi-SFDAGR} addresses cross-domain issues using Unsupervised Domain Adaptation (UDA) for unlabeled test data scenarios

\section{Method}

\begin{figure*}[t] 
  \centering
  \includegraphics[width=\textwidth]{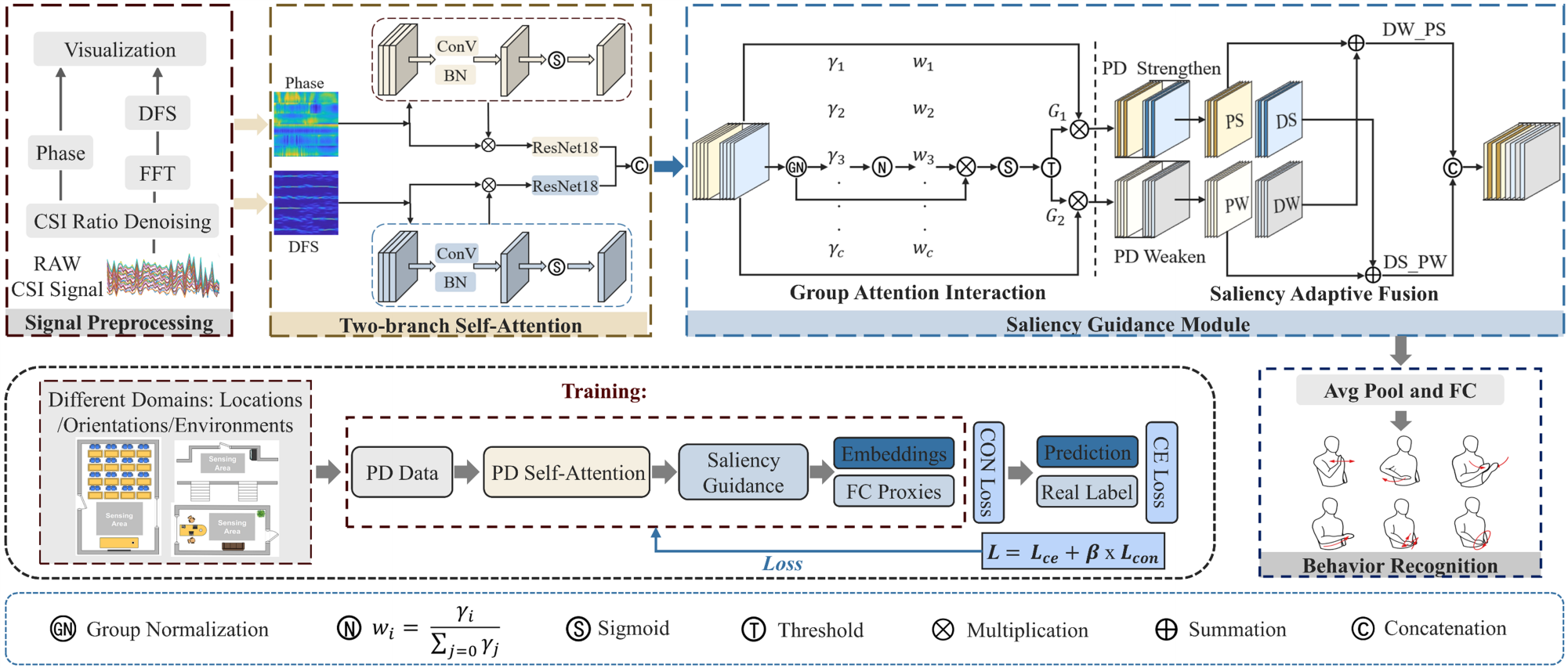} 
  \caption{Framework Overview of Wi-CBR. The proposed Wi-CBR framework integrates Signal Preprocessing, a Two-branch Self-attention Module, and a Saliency Guidance Module to achieve robust behavior recognition across domains. It leverages phase information for temporal dynamics and Doppler Frequency Shift (DFS) for spatial motion characteristics. A Group Attention Interaction mechanism identifies salient features, while Saliency Adaptive Fusion effectively combines critical and non-critical behavior features. The system is designed to generalize across different environments, orientations, and locations, ensuring superior performance in both in-domain and cross-domain scenarios.}
  \label{fig:The System framework of Wi-CBR.}
\end{figure*}

As shown in Fig. \ref{fig:The System framework of Wi-CBR.}, Wi-CBR consists of four components: signal preprocessing, two-branch self-attention learning, saliency guidance, and classification prediction. In signal preprocessing, the CSI-ratio model handles denoising, while DFS is extracted using STFT. The phase and DFS matrices are visualized as images for deep learning. The network employs two-branch self-attention and pre-trained ResNet18 for initial feature extraction, followed by feature fusion via a cross-model interactive module. Finally, behavior prediction is achieved using a classifier with dual-loss constraints.

\subsection{Task Definition}
WiFi CSI describes the signal's attenuation on its propagation paths, such as scattering, multipath fading or shadowing, and power decay over distance. It can be characterized as:
\begin{equation}
\mathbf{Y} = \mathbf{H} \cdot \mathbf{X} + \mathbf{N},
\end{equation}
where $\mathbf{Y}$ and $\mathbf{X}$ are the received and transmitted signal vectors, respectively. $\mathbf{N}$ is the additive white Gaussian noise, and $\mathbf{H}$ is the channel matrix representing CSI. CSI is a superposition of signals of all the propagation paths, and its channel frequency response (CFR) can be represented as:
\begin{equation}
H(f, t) = \sum_{m \in \Phi} a_m(f, t)e^{-j2\pi \frac{d_m(t)}{\lambda}},
\end{equation}
where $f$ and $t$ represent center frequency and time stamp, $m$ is the multipath component. $a_m(f, t)$ and $d_m(t)$ denote the complex attenuation and propagation length of the $m$th multipath component, respectively. $\Phi$ denotes the set of multipath components, and $\lambda$ is the signal wavelength.
In the case of CSI-based gesture recognition, the multipath component $m$ consists of dynamic and static paths:
\begin{align}
H(f, t) &= \sum_{m_s \in \Phi_s} a_{m_s}(f, t) e^{-j2\pi \frac{d_{m_s}(t)}{\lambda}} \notag\\
        &+ \sum_{m_d \in \Phi_d} a_{m_d}(f, t) e^{-j2\pi \frac{d_{m_d}(t)}{\lambda}} \label{eq:H}
\end{align}

\subsection{CSI Denoising Preprocessing}
As demonstrated in the previous section, the gesture can be portrayed by the change of phase shift in CSI. Unfortunately, for commodity WiFi devices, as the transmitter and receiver are not synchronized, there exists a time-varying random phase offset $e^{-j\theta_{\text{offset}}}$ :
\begin{align}
H(f, t) 
&= e^{-j\theta_{\text{offset}}} ( a_{m_s}(f, t) e^{-j2\pi \frac{d_{m_s}(t)}{\lambda}}   \notag\\
&+ a_{m_d}(f, t) e^{-j2\pi \frac{d_{m_d}(t)}{\lambda}} ) ,\label{eq:H1}
\end{align}
where $e^{-j2\pi \frac{d(t)}{\lambda}}$, and $d(t)$ denote phase-shift, and path length of dynamic components, respectively. This random offset thus prevents us from directly using the CSI phase information.

Therefore, we need to eliminate $e^{-j\theta_{\text{offset}}}$. Fortunately, for commodity WiFi cards, this random offset remains the same across different antennas on the same WiFi network interface card (NIC) as they share the same RF oscillator. Thus, it can be eliminated by the CSI-ratio model:
\begin{equation}
H_r(f, t) = \frac{H_1(f, t)}{H_2(f, t)}
= \frac{e^{-j\theta_{\text{offset}}} (H_{s,1} + H_{d,1})}{e^{-j\theta_{\text{offset}}}(H_{s,2} + H_{d,2})} 
\end{equation}
where $H_1(f, t)$ and $H_2(f, t)$ are the CSI of two receiving antennas. When two antennas are close to each other, $\Delta d$ can be regarded as a constant. According to Mobius transformation ~\cite{csi1}, (5) represents transformations such as scaling and rotation of the phase-shift $e^{-j2\pi \frac{d_1(t)}{\lambda}}$ of antenna 1 in the complex plane, and these transformations will not affect the changing trend of the phase-shift\cite{csi2,csi3,csi4}.

\textbf{CSI to Phase.}
Phase ratio $ \text{P} $ extracted from $ H_r $ can be used to describe gestures:
\begin{equation}
\text{P} = angle(H_r), 
\end{equation}
where $angle(\cdot)$ denotes the phase extraction function. For a complex $ z = \text{abs}(z) \cdot e^{j\theta} $, we can use $angle(\cdot)$ to obtain the phase of $ z $, $ \theta = angle(z) $.

\textbf{CSI to DFS.}
To extract Doppler Frequency Shift (DFS) features, we take the raw CSI signal received by a specific  Tx–Rx antenna pair, denoted as $H_q(f, t)$, as an example. This stream contains temporal variations caused by human motion. Unlike $H_r(f, t)$, which is a ratio of two antenna CSI streams designed to remove phase offset for phase analysis, $H_q(f, t)$ retains the original frequency content, making it more suitable for Doppler-based gesture analysis. The DFS reflects frequency changes due to gesture motion, such as hand speed or direction. 

\begin{equation}
S_q(\tau, \omega) = \int_{-\infty}^{\infty} H_q(f, t) w(t - \tau) e^{-j \omega t} dt\ ,
\end{equation}
where $w(t - \tau)$ is a window function (e.g., Hanning window) to segment the signal, $\tau$ is the time localization, and $\omega$ is the frequency in radians, tied to the Doppler shift. The power spectrogram is then computed:
\begin{equation}
\text{D} = |S_q(\tau, \omega)|^2 = \left| \int_{-\infty}^{\infty} H_q(f, t) w(t - \tau) e^{-j \omega t} dt \right|^2\ 
\end{equation}

In this spectrogram, the frequency $\omega$ at each time $\tau$ corresponds to the Doppler shift $f_d = \omega / (2\pi)$. $f_d$ is proportional to the gesture speed $v(t)$. Positive frequencies indicate motion toward the receiver, while negative frequencies indicate motion away. This time-frequency representation enables detailed analysis of gesture dynamics. 

For the Widar3.0 dataset, each DFS file is a $6 \times 121 \times \mathrm{T}$ matrix, where the first dimension represents the 6 receivers, the second dimension represents the 121 frequency segments ranging from $[-60, 60]$ Hz, and the third dimension represents the timestamps with a sampling rate of 1000 Hz. 

\subsection{Two-branch Self-attention Learning}

\textbf{Two-branch Self-attention.}
The proposed dual-path spatial attention mechanism processes phase ($\mathbf{P}_x$) and Doppler ($\mathbf{D}_x$) features through independent attention branches. Let $f_{\text{conv}}^{(k)}$ denote a convolution operation with kernel size $k \times k$, and $\text{BN}$ represent batch normalization. The attention weights are computed as:
\begin{align}
    \mathbf{A}_p &= \sigma\left(\text{BN}\left(f_{\text{conv}}^{(7)}\left(f_{\text{conv}}^{(7)}(\mathbf{P}_x)\right)\right)\right).
\end{align}
where $\sigma(\cdot)$ is the sigmoid function. $\mathbf{A}_d$ is computed in the same way using $\mathbf{D}_x$ as input. The refined features are obtained through:
\begin{equation}
    \begin{alignedat}{2}
        \mathbf{P}_{\text{out}} &= \mathbf{P}_x \otimes \mathbf{A}_p \oplus \mathbf{P}_x; &\quad 
        \mathbf{D}_{\text{out}} &= \mathbf{D}_x \otimes \mathbf{A}_d \oplus \mathbf{D}_x 
    \end{alignedat}
\end{equation}
where $\otimes$ is element-wise multiplication, $\oplus$ is element-wise summation. Each branch maintains independent convolution parameters, with channel dimensions preserved in the first convolution ($3 \rightarrow 3$) and reduced to a single channel ($3 \rightarrow 1$) in the second convolution.

\textbf{Two-branch Feature Extraction.} Independent spatial attention parameters for P/D branches; Separate batch normalization statistics for each modality;
Symmetric padding (3 pixels) maintained in all convolutions; No parameter sharing between $\phi_{\text{ResNet}}^P$ and $\phi_{\text{ResNet}}^D$.

\begin{equation}
    \mathbf{X}_{\text{PD}} = \mathrm{Concat}(\phi_{\text{ResNet}}^P(\mathbf{P}_{\text{out}}), \phi_{\text{ResNet}}^D(\mathbf{D}_{\text{out}})) \in \mathbb{R}^{1024 \times 7 \times 7}
\end{equation}

\subsection{Saliency Guidance Module}

\textbf{Group Attention Interaction.} To make the extracted Phase and DFS features interact and merge, we use group normalization to achieve the attention of different channels. The generated attention maps are thresholded to get the strengthened and weakened maps, and then the strengthened and weakened Phase and DFS features are obtained. Based on retaining important features and attenuating minor features ~\cite{scconv}, the feature space redundancy is reduced while utilizing both features.

Separate operation aims to separate those informative feature maps from less informative ones corresponding to the spatial content. We leverage the scaling factors in Group Normalization (GN) ~\cite{GN} layers to assess the informative content of different feature maps. To be concrete, given an intermediate feature map $X_{\text{PD}}  \in \mathbb{R}^{N\times C\times H\times W}$, where $N$ is the batch axis, $C$ is the channel axis, $H$ and $W$ are the spatial height and width axes. We first standardize the input feature $X$ by subtracting mean $\mu$ and dividing by standard deviation $\sigma$ as follows:
\begin{equation}
X_{\text{gn}} = \mathrm{GN}(\mathbf{X}_{\text{PD}}) = \gamma \frac{X_{\text{PD}} - \mu}{\sqrt{\sigma^2 + \epsilon}} + z 
\end{equation}
where $\mu$ and $\sigma$ are the mean and standard deviation of $X_{\text{PD}}$, $\epsilon$ is a small positive constant added for numerical stability, $\gamma$ and $z$ are trainable affine transformation parameters.

Noted that we leverage the trainable parameters $\gamma \in \mathbb{R}^C$ in GN layers as a way to measure the variance of spatial pixels for phase and DFS. The richer spatial information reflects more variation in spatial pixels contributing to a larger $\gamma$. The normalized correlation weights $W_\gamma \in \mathbb{R}^C$ are obtained by equation \eqref{eq:14}, which indicates the importance of different phase and DFS feature maps.

\begin{equation}
\label{eq:13}
W_\gamma = \{w_i\} = \frac{\gamma_i}{\sum_{j=1}^C \gamma_j}, \quad i, j = 1, 2, \dots, C 
\end{equation}
Then the weight values of feature maps reweighted by $W_\gamma$ are mapped to the range (0, 1) by the sigmoid function and gated by a threshold. We set those weights above the threshold to 1 to obtain the informative weights $G_1$ while setting them to 0 to gain the non-informative weights $G_2$ (the threshold is set to 0.5 in the experiments):
\begin{equation}
\label{eq:14}
G = \text{Gate}(\text{Sigmoid}(W_\gamma(GN(X_{\text{PD}}))))
\end{equation}

We multiply input features $X_{\text{PD}}$ by $G_1$ and $G_2$ respectively, yielding two weighted features: the strengthen  ones $X_{\text{PD}}^S$ and less informative ones $X_{\text{PD}}^W$.Thus, we completed the interaction between the input phase and DFS features. The attention weights are obtained by learning the variance and bias through group normalization, which is gated to obtain the strengthened and weakened attention maps. Two components obtained are as follows: $X_{\text{PD}}^S$ has informative and expressive spatial content and is strengthened, while $X_{\text{PD}}^W$ has little or no information, which is considered redundant and weakened.

\textbf{Saliency Adaptive Fusion.}
We propose a fusion operation to achieve synergistic utilization. An information-rich feature after strengthening is added to a feature with less information after weakening. New features with richer information are generated, i.e., one feature dominates while the other aids. Instead of adding the two components directly, we use a cross-fertilization operation to thoroughly combine the two weighted different information features to enhance the information flow between them. The cross-rendered features $Y_{\text{PS\_DW}}$ and $Y_{\text{PW\_DS}}$ are then stitched together to obtain a spatially fine feature mapping of $X_{\text{out}}$.
The whole process of Fusion operation can be expressed as :
\begin{equation}
\begin{cases}
\label{eq:15}
X_{\text{PD}}^S = G_1 \otimes X_{\text{PD}},
X_{\text{PD}}^W = G_2 \otimes X_{\text{PD}}; \\
Y_{\text{DS\_PW}} = X_{\text{D}}^S \oplus X_{\text{P}}^W,
Y_{\text{DW\_PS}} = X_{\text{D}}^W \oplus X_{\text{P}}^S;\\
X_{\text{out}} = \mathrm{Concat}(Y_{\text{DS\_PW}}, Y_{\text{DW\_PS}})
\end{cases}
\end{equation}
where $\otimes$ is element-wise multiplication, $\oplus$ is element-wise summation, $\mathrm{Concat}$ is concatenation.
When DFS is remarkable as a domain-independent feature, we use the enhanced DFS feature, assisted by the weakened detailed phase feature, namely $Y_{\text{DS\_PW}}$.
When DFS is weak as a domain-independent feature, we use the weakened DFS feature and use the enhanced detailed phase feature as a supplement, that is, $Y_{\text{DW\_PS}}$.
After the Saliency Guidance Module is applied to the input features $X_{\text{PD}}$, not only do we separate the informative features from less informative ones, but also we reconstruct them to enhance the representative features and suppress the redundant features in spatial dimension. 

\subsection{Contrastive Loss Optimization}
Building upon the aforementioned modules, we obtain the Cross-Model Fusion feature representation $X_{\text{out}}$. Most existing works directly feed the global representation $X_{\text{out}}$ into the classifier (i.e., a fully-connected layer with a Softmax) to predict the probability of gestures $\hat{y}$. The model is then trained by minimizing the corresponding loss $L_{ce}$ between the prediction values $\hat{y}$ and their ground truths $y$.

\begin{equation}
\mathcal{L}_{ce} = \mathcal{L}_{ce}(y, \hat{y}) = -\frac{1}{M} \sum_{m=1}^{M} \sum_{s=1}^{S} y_{m,s} \log(\hat{y}_{m,s})
\end{equation}
where $L_{ce}$ is a classification cross-entropy loss, and $M$ is the number of data samples, $S$ is the number of gestures.

\textbf{Contrastive Loss.} optimizes the objective by learning a distance measure based on multiple positive and negative sample-to-sample pairs. The key idea behind this is to learn an embedding space where similar pairs of samples are close to each other and dissimilar pairs are far apart. Thus, we can obtain an invariant representation across different environments for the same instance, i.e., a domain-independent characterization. 
Several elements can influence the variation patterns of the CSI signal, with varying degrees of effect. For example, a user's influence on signal patterns is less significant compared to changes in location and orientation. Importantly, positive sample pairs across diverse environments differ, and some may be challenging to match due to substantial data discrepancies. Perfectly aligning all samples might limit the model's ability to generalize. To mitigate this, we utilize class proxies to symbolize each gesture, ideally enhancing resilience across samples from varied settings. Formally, these proxy vectors are defined as the weights of the final fully connected layer in the classifier. To further improve semantic consistency, we implement a \textbf{proxy-based} contrastive loss that utilizes the connections between class proxies and samples to foster robust representations ~\cite{Proxy}.
Given the representation $x_i$ of the $i$th sample, we select its class proxy $w_c$ in place of positive samples $x_+$ to form proxy-to-sample positive pairs. The contrast loss is incorporated into the overall loss function \( \mathcal{L} \):
\begin{align}
\mathcal{L}_{con} 
&= -\frac{1}{N} \sum_{i=1}^{N} \log \frac{e^{(w_c^T xi)/\tau}}{e^{(w_c^T x_i)/\tau} + \sum_{k=1, k \neq c}^{R} e^{(w_k^T x_i)/\tau}} ;\notag\\
&\mathcal{L} = \mathcal{L}_{ce} + \beta \times \mathcal{L}_{con} \label{eq:H3}
\end{align}
where $w_c$ represents the class proxy corresponding to class $c$, $R$ is the total number of classes, and \(\tau\) serves as the temperature parameter. We aim to minimize the following final loss function, where $\mathcal{L}_{ce}$ is the cross-entropy loss and $\mathcal{L}_{con}$ is the proxy-based contrastive loss. $\beta$ is the trade-off parameter.

\begin{table*}[h]
  \centering  
  \begin{tabular}{l l c c c c c}
    \toprule
    \multirow{2}{*}{Method} & 
    \multirow{2}{*}{Processing Flow} & 
    \multicolumn{3}{c}{Widar3.0} & 
    \multirow{2}{*}{Mean} & 
    \multirow{2}{*}{XRF55} \\
    \cmidrule(lr){3-5}
    & & \multicolumn{1}{c}{CL} & \multicolumn{1}{c}{CO} & \multicolumn{1}{c}{CE} & & \\
    \midrule
    EI~\cite{EI}       & CSI$\rightarrow$Amplitude & 73.33 & 79.70 & {--}   & {--} & {--} \\
    Widar3.0~\cite{Widar3.0} & CSI$\rightarrow$DFS$\rightarrow$BVP & 90.48 & 81.58 & 83.30 & 85.12 & {--} \\
    WiHF~\cite{WIHF}      & CSI$\rightarrow$DFS$\rightarrow$MCP & 91.22 & 80.64 & {--} & {--} & {--} \\
    THAT~\cite{THAT}     & Raw CSI & 71.56 & 81.76 & 49.71 & 67.68 & 23.23 \\
    WIGRUNT~\cite{WIGRUNT}  & CSI$\rightarrow$Phase & 97.08 & 93.39 & 95.36 & 95.28 & 55.92 \\
    AaD~\cite{AaD}      & CSI$\rightarrow$Phase & 95.90 & 95.38 & 93.00 & 94.83 & 55.64 \\
    ImgFi~\cite{Imgfi}    & CSI$\rightarrow$STFT, RT image & 39.58 & 38.12 & 40.37 & 39.36 & 31.90 \\
    WiSR~\cite{WISR}      & CSI image & 67.73 & 69.74 & 52.77 & 63.41 & 26.66 \\
    Recurrent ConFormer~\cite{Recurrent_ConFormer} & Raw CSI & 73.84 & 85.88 & 50.38 & 70.03 & 16.54 \\
    WiDual~\cite{WiDual}    & CSI$\rightarrow$Phase & 97.39 & 94.87 & {--} & {--} & {--} \\
    WIGNN~\cite{WiGNN}    & CSI$\rightarrow$DFS & 95.20 & 93.30 & {--} & {--} & {--} \\
   
    Wi-SFDAGR~\cite{Wi-SFDAGR} & CSI$\rightarrow$Phase & 97.30 & \textbf{97.17} & 95.52 & 96.66 & 57.99 \\
  
 \textbf{Wi-CBR (Ours)}    & \textbf{CSI$\rightarrow$Phase}, \textbf{DFS} & \textbf{98.34} & 96.30 & \textbf{96.87} & \textbf{97.17} & \textbf{66.05} \\
    \bottomrule
  \end{tabular}
\caption{THE accuracy of Wi-CBR under CL(Cross-Location), CO(Cross-Orientation), and CE(Cross-Environment) settings in the WIDAR3.0 dataset and the XRF55 dataset.}
\label{table1}
\end{table*}
\section{Experiments}

\subsection{Experimental Setup}
We train and evaluate our proposed Wi-CBR on the two largest human sensing multimodal public datasets.

\textbf{Gesture Recognition.} To evaluate the effectiveness of our model for cross-domain gesture recognition, we conducted extensive experiments on the Widar 3.0 and XRF55 datasets. For in-domain, cross-location, and orientation evaluations on Widar 3.0, we used 80\% of the data for training and 20\% for testing with five-fold cross-validation. For cross-location evaluation, one location was used for testing and the remaining four for training. The in-domain and cross-direction evaluations followed a similar approach. For cross-environment evaluation, we used data from three environments, totaling 12,750 samples, with training on two environments and testing on the third using triple cross-validation. For XRF55, the cross-environment evaluation involved four scenarios, with quad-fold cross-validation on 6,240 samples.

\textbf{Activity Recognition.} For the fairness of the experimental evaluation, the implementation details are exactly the same as those of XRF55. Note that the quad-fold cross-validation is not used here. At this time, due to the lack of multi-environment data support, the cross-domain migration capability of the model is highly required. The training set of scene 1 is used for training, that is, the first 14 times of each behaviour of each user. The test set is tested with the samples of scenes 2, 3, and 4, respectively. There are 33000 samples in the training set and 3300 samples in each scene in the test set, including 55 daily human behaviours.

\textbf{Implementation Details.}
We used MATLAB to preprocess CSI data and generate 224×224 RGB images. After obtaining phase and DFS images, all models were implemented in PyTorch 1.13.1. The network architecture and data dimensions are shown in Fig. \ref{fig:The System framework of Wi-CBR.}. Wi-CBR employs ResNet-18~\cite{60} with pre-trained ImageNet weights as the feature extractor. The Cross-Model Interaction module set the threshold to 0.5. The contrast loss weight $\beta$ and temperature are both set to 0.1. During training, the model is optimized using Adam with a learning rate of 0.0001, a batch size of 10, and 30 epochs. We used the same network structure for both the Widar 3.0 and XRF55 datasets, ensuring robustness across datasets. Full connection layer classification header: in gesture recognition, it is 6 to 9 in widar3.0 dataset, 8 in XRF55 dataset, and 55 in activity recognition. A random seed of 42 was set for reproducibility.

\subsection{Comparisons to Prior SOTA Results}
We compare our method with the baseline method in downstream human perception tasks, including fine-grained HGR and human HAR. We mainly study single-factor cross-domain, that is, only one domain factor we have never seen in location, orientation, and environment. We trained in some specific domains and test in an unknown domain.

\begin{table}[t]
\centering
\resizebox{0.48\textwidth}{!}{   
\begin{tabular}{@{}ccccc@{}}
\toprule
\multicolumn{1}{c}{\multirow{2}{*}{CE}} & \multicolumn{1}{c}{\multirow{2}{*}{Method}} & \multicolumn{3}{c}{few-shot} \\
\cmidrule(lr){3-5}
 & & zero & one & two \\
\midrule
\multirow{3}{*}{2} 
    & WIGRUNT\cite{WIGRUNT}  & 21.82 & 41.34 & 50.17 \\
    & XRF55\cite{XRF55}    & 2.52  & 49.83 & 57.51 \\
    & \textbf{Wi-CBR (Ours)} & \textbf{31.33} & \textbf{50.92} & \textbf{58.96} \\
\midrule
\multirow{3}{*}{3} 
    & WIGRUNT\cite{WIGRUNT}  & 13.12 & 41.34 & 56.53 \\
    & XRF55\cite{XRF55}    & 2.14  & 50.85 & \textbf{63.30} \\
    & \textbf{Wi-CBR (Ours)} & \textbf{28.60} & \textbf{51.74} & 61.01 \\
\midrule
\multirow{3}{*}{4} 
    & WIGRUNT\cite{WIGRUNT}  & 14.55 & 42.60 & 53.81 \\
    & XRF55\cite{XRF55}    & 2.03  & 50.94 & 61.41 \\
    & \textbf{Wi-CBR (Ours)} & \textbf{31.30} & \textbf{52.70} & \textbf{62.66} \\
\bottomrule
\end{tabular}
}
\caption{55 kinds of human daily behaviors cross-domain recognition on XRF55 under few-shot.}
\label{table2}
\end{table}

\begin{figure*}[h]
  \centering
  \includegraphics[width=\textwidth]{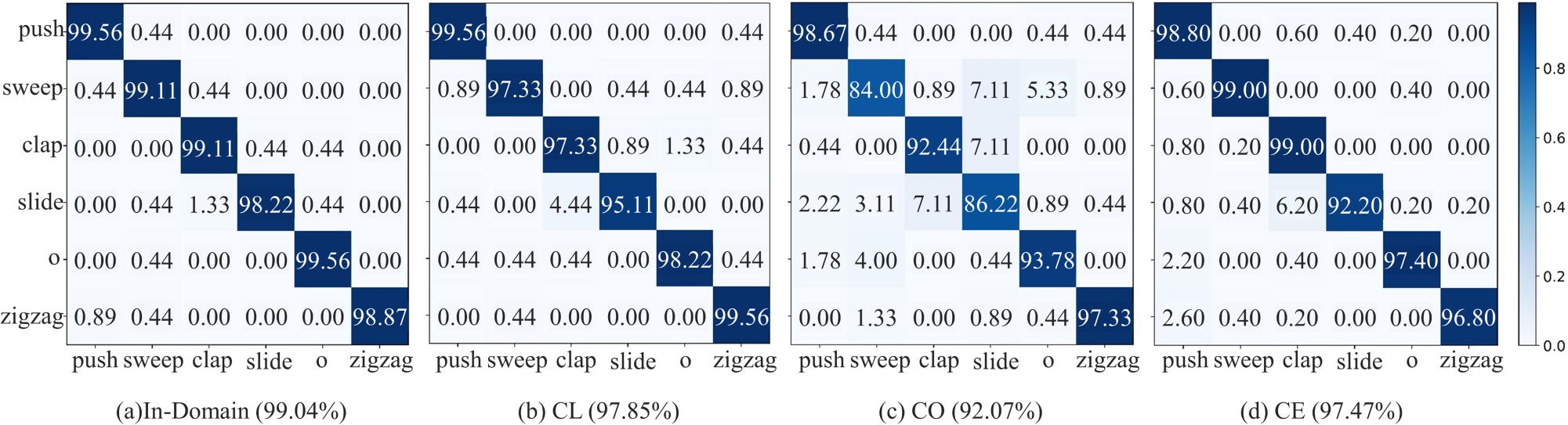} 
  \caption{The confusion matrices in the ID (repetition 1), CL (location 1), CO (orientation 1), and CE (environment 3).}
  \label{fig:fig4}
\end{figure*}

\textbf{HGR Comparison with SOTA Methods.}
We compare the cross-domain performance with SOTA methods across modeling-based, low-level semantic, and high-level semantic approaches (Tab. \ref{table1}). Modeling-based methods (e.g., Widar3.0~\cite{Widar3.0}, WiHF~\cite{WIHF}) extract DFS and domain-independent features, achieving 85.12\% on Widar3.0. Low-level semantic methods (e.g., ImgFi~\cite{Imgfi}, WiSR~\cite{WISR}) rely on raw CSI or visualizations, with performance ranging from 40\% to 70\% on Widar3.0 and up to 31.90\% on XRF55. High-level semantic methods (e.g., Wi-SFDAGR~\cite{Wi-SFDAGR}) utilize CSI-ratio denoising and advanced networks like ResNet18, reaching 96.66\% on Widar3.0 through optimized feature aggregation.

Wi-CBR outperforms state-of-the-art on both datasets. In addition, we observe an interesting phenomenon: The performance across environments on the XRF55 dataset is lower than that of Widar3.0, regardless of which method is used. This is because the XRF55 dataset has a lower sampling rate than Widar3.0, and the number of receiver arrangements in the environment is half that of Widar3.0. Wi-CBR, on the other hand, demonstrates robustness across datasets, and our work still achieves 66.05\% accuracy despite changes in conditions such as receiver and sampling rate.

\begin{table}[t]
  \centering
  \begin{tabular}{c c c c c}
\toprule
    Methods & ID & CL & CO & CE \\
    \midrule
    Without DFS            & 99.42 & 96.07 & 92.76 & 93.76 \\
    Without Phase          & 97.13 & 95.04 & 87.09 & 94.93 \\
    Without CL             & 99.49 & 97.26 & 95.56 & 96.08 \\
    \bottomrule
  \end{tabular}
\caption{Effectiveness of each component.}
\label{table3}
\end{table}

\textbf{HAR included Comparison on XRF55.}
We further evaluate the recognition of human daily activities. We initially train the model using training samples from environment 1, and
subsequently finetune it using 0/1/2 samples (per subject per action) from environment 2. We then test the finetuned
model with the remaining 20/19/18 samples (per subject per action) in environment 2. Tab. \ref{table2} indicates the model’s best transferability to unseen environment.
In the case of zero-shot, most models can not effectively learn the domain invariance feature, and the model is close to a random guess. The advantage of Wi-CBR is that it can recognize the good features learned by multimodal collaborative sensing even if it is only learned in one environment, without using any samples of an unknown environment.

\subsection{Ablation Study} 
We evaluated the impact of different components and fusion methods on the experiment.
The results in Tab. \ref{table3} show that DFS greatly improves cross-environment performance, while phase mainly affects in-domain and cross-domain performance within the environment. Additionally, omitting contrast loss guidance led to a noticeable drop in cross-domain performance, while in-domain performance remained stable, indicating that contrast loss helps the model focus on gesture-related features rather than environmental noise.
The results in Tab.~\ref{table4} show that
easy channel attention using WiGRUNT ~\cite{WIGRUNT} channel-gate module will continue to reduce the accuracy of the model, highlighting the importance of interaction for balancing phase and DFS data. We also tried different fusion methods. In fact, when there is only one kind of data as input, it is equivalent to self-interaction, such as using \textbf{PS\_PW}. When using the two kinds of data as input, the respective enhanced addition \textbf{PS\_DS} exists feature redundancy and cannot play a complementary effect.  Overall, Wi-CBR effectively leverages full CSI raw data and DFS for improved cross-domain performance.

\begin{table}[t]
  \centering
  \begin{tabular}{c c c c c}
\toprule
    Methods & ID & CL & CO & CE \\
    \midrule
    Channel Attention  & 99.11 & 97.07 & 95.07 & 95.93 \\
    PS\_DS and PW\_DW  & 99.54 & 97.14 & 95.13 & 96.28 \\
    \textbf{Wi-CBR}    & \textbf{99.54} & \textbf{98.34} & \textbf{96.30} & \textbf{96.57} \\
    \bottomrule
  \end{tabular}
\caption{Effectiveness of Fusion.}
\label{table4}
\end{table}

To analyze recognition accuracy and misclassification rates, we present confusion matrices for test location 1, orientation 1, and environment 3 on Widar 3.0, as shown in Fig. \ref{fig:fig4}.
In cross-location and cross-orientation tests, “push” and “zigzag”  achieve the best accuracy, while “slide” and “sweep” show the lowest, respectively. In the cross-environment test, “sweep” and “clap” perform best. 
Interestingly, “slide” is frequently misclassified as “clap.” This likely stems from similar motion trajectories, as both gestures involve inclined or sliding movements, making them harder to distinguish.

\section{Conclusion}
In this work, we propose Wi-CBR, a WiFi-based cross-domain human behavior recognition framework with strong cross-domain performance. It integrates model-based and learning-based methods, using phase data (path length changes) and DFS data (frequency shifts) for robust cross-domain performance. It employs a two-branch self-attention module to extract spatio-temporal features and a saliency-aware adaptive mechanism to enhance phase features based on DFS significance. This time-frequency fusion focuses on behavior while reducing environmental interference, improving recognition accuracy.

\section{Acknowledgments}
This work was supported by the National Natural Science Foundation of China (Grants No. 62502144, 61932009, 62462015), the Anhui Provincial Natural Science Foundation, China (Grant No. 2408085QF191), the Young Elite Scientist Sponsorship Program by Gast (Grant No. GASTYESS202429), and the Fundamental Research Funds for the Central Universities (Grants No. JZ2024HGTA0178, PA2025IISL0112). The computation is completed on the HPC Platform of Hefei University of Technology. 

\bibliography{references}

\clearpage

\twocolumn[{
\begin{center}
{\LARGE\bfseries Appendix}
\end{center}
\vspace{0.5em}
}]

\appendix

\section{A. More Experimental Details}
\subsection{A.1 More Experiment Results}
For the Widar 3.0 dataset, Wi-CBR has an average recognition accuracy of 99.54\% for in-domain gesture recognition and 98.34\%, 96.30\%, and 96.87\% for cross location, orientation, and environment, respectively. For the XRF55 dataset, the average recognition accuracy across environments is 66.05\%. The detailed results are shown in Tab. \ref{table7}. The Setting denotes the data for testing, while others are for training. We report the results on accuracy and macro F1 with additional random seeds as Tab. \ref{tab:seed-results}.

\begin{table}[h]
\centering
\small  
\setlength{\tabcolsep}{4pt}  
\begin{tabular}{lcccccc}
\toprule
Dataset & Setting & 1 & 2 & 3 & 4 & 5 \\
\midrule
\multirow{4}{*}{Widar 3.0} 
    & In-Domain & 99.04 & 99.70 & 99.48 & 99.78 & 99.70 \\
    & CL & 97.85 & 97.63 & 98.74 & 98.00 & 99.48 \\
    & CO & 92.07 & 98.22 & 97.85 & 99.19 & 94.15 \\
    & CE & 93.04 & 99.40 & 98.16 & \texttt{-} & \texttt{-} \\
\cmidrule(lr){2-7}    
XRF55 & CE & 55.65 & 68.13 & 72.50 & 67.92 & \texttt{-} \\
\bottomrule
\end{tabular}
\caption{Detailed gesture recognition performance on Widar 3.0 and XRF55.}
\label{table7}
\vspace{-5mm}
\end{table}

\begin{table}[h]
\centering
\setlength{\tabcolsep}{4pt}
\begin{tabular}{lccccc}
\toprule
Seed & 21 & 888 & 1025 & 2024 & Avg$\pm$Std \\
\midrule
Widar3.0-Acc & 97.21 & 96.92 & 96.79 & 96.59 & 96.88$\pm$0.23 \\
Widar3.0-F1  & 97.12 & 96.90 & 96.77 & 96.73 & 96.88$\pm$0.23 \\
XRF55-Acc    & 66.69 & 67.20 & 64.44 & 65.15 & 65.80$\pm$1.13 \\
XRF55-F1     & 66.50 & 66.70 & 64.00 & 64.80 & 65.25$\pm$1.12 \\
\bottomrule
\end{tabular}
\caption{Results under different random seeds.}
\label{tab:seed-results}
\vspace{-5mm}
\end{table}

\subsection{A.2 Impact of number of users and gestures}
We evaluate Wi-CBR under varying numbers of users (default: 9) and gestures (default: 6), with results in Tab.~\ref{tab:wigrunt_wicbr_users} and Tab.~\ref{tab:wigrunt_wicbr_gestures}, respectively. 
As users increase, Wi-CBR generally improves , indicating better adaptation to more users, whereas WiGRUNT declines, with a sharp drop when adding the first extra user; overall, Wi-CBR benefits from more users. 
As gestures increase, both methods degrade; however, Wi-CBR drops more slowly than WiGRUNT for in-domain and cross-location and can rebound on easily recognized gestures, while both show similar trends under cross-orientation. 
WiGRUNT decreases monotonically, suggesting Wi-CBR is more stable as gesture classes grow.

\begin{table}[h]
\centering
\small
\setlength{\tabcolsep}{4pt}  
\begin{tabular}{llcccc}
\toprule
\multicolumn{2}{c}{No. of Users} & 6 & 7 & 8 & 9 \\
\midrule
\multirow{3}{*}{WiGRUNT}
& In-Domain & 99.60 & 99.45 & 99.48 & 99.10 \\
& CL         & 96.49 & 96.21 & 96.72 & 96.17 \\
& CO         & 93.36 & 92.86 & 94.11 & 93.95 \\
\midrule
\multirow{3}{*}{Wi-CBR}
& In-Domain & 99.07 & 99.54 & 99.63 & \textbf{99.54} \\
& CL         & 97.20 & 98.46 & 98.92 & \textbf{98.34} \\
& CO         & 95.31 & 96.84 & 97.08 & \textbf{96.57} \\
\bottomrule
\end{tabular}
\caption{Accuracy of WiGRUNT and Wi-CBR for different numbers of users.}
\label{tab:wigrunt_wicbr_users}
\vspace{-5mm}
\end{table}

\begin{table}[h]
\centering
\small  
\setlength{\tabcolsep}{4pt} 
\begin{tabular}{llcccc}
\toprule
\multicolumn{2}{c}{No. of Gestures} & 6 & 7 & 8 & 9 \\
\midrule
\multirow{3}{*}{WiGRUNT}
& In-Domain & \textbf{99.60} & 99.26 & 98.32 & 98.08 \\
& CL         & 96.49 & 95.45 & 92.25 & 93.58 \\
& CO         & 93.36 & 92.12 & 89.22 & 89.01 \\
\midrule
\multirow{3}{*}{Wi-CBR}
& In-Domain & 99.07 & 99.31 & 98.67 & 98.61 \\
& CL         & \textbf{97.20} & 97.52 & 95.77 & 96.03 \\
& CO         & \textbf{95.31} & 94.89 & 91.20 & 91.11 \\
\bottomrule
\end{tabular}
\caption{Accuracy of WiGRUNT and Wi-CBR for different numbers of gestures.}
\label{tab:wigrunt_wicbr_gestures}
\vspace{-5mm}
\end{table}

\section{B. More Dataset Details}

\subsection{B.1 Gesture Recognition Dataset Details}

\textbf{Widar3.0} is one of the most widely used datasets
for WiFi-based gesture recognition. The dataset collection device consists of one transmitter and six receivers, each equipped with an Intel 5300 wireless card. Each receiver has three antennas arranged in a line. The operating frequency
of the device is 5.825 GHz, with a sampling rate of 1000 packets per second. Widar3.0 mainly consists of two types of
datasets. Firstly, there are 12,000 gesture samples commonly used in human-computer interaction (16 users × 5 locations ×
5 orientations × 6 gestures × 5 instances). Secondly, there are 5,000 gesture samples representing digits 0-9 on a horizontal plane (2 users × 5 locations × 5 orientations × 10 gestures × 5 instances). To ensure a fair comparison, we only utilized 4,500 gesture samples (9 users × 5 locations × 5 orientations × 6 gestures × 5 instances). The detailed descriptions of Widar3.0 dataset are provided in Tab. \ref{tab:widar-dataset}.

\textbf{XRF55} is a large multi-radio frequency dataset forhuman action analysis. To verify the effectiveness of the proposed Wi-CBR, we selected only the samples related to gesture actions collected using WiFi devices (e.g., drawing a circle,
drawing a cross, pushing, pulling, swiping left, swiping right,swiping up, and swiping down) for the experiments. Among
them, 30 users in environment 1 repeated the gestures 20 times within 5 seconds, and the remaining 9 users repeated
the gestures 20 times within 5 seconds in the other three environments, totaling 6,240 samples.The detailed descriptions of XRF55 dataset are provided in Tab. \ref{tab:xrf55-dataset}.

\subsection{B.2 Activity Recognition Dataset Details}

XRF55 includes 15 human-object interaction actions, 7 human-human interaction actions, 8 fitness actions,
14 body motion actions, and 11 human-computer instruction actions. These actions were meticulously selected
from 19 RF sensing papers and 16 video action recognition datasets. Each action is chosen to support various
applications of high practical value, such as elderly fall detection, fatigue monitoring, domestic violence detection,and even accommodating scenarios where one’s hands might be preoccupied or unavailable, like individuals with certain disabilities or temporary injuries. These actions are likely to attract substantial attention from the industrial and academic sectors of the RF sensing community.The detailed activities are shown in Tab. \ref{tab:XRF55}.

\begin{table*}[h]
\centering
\begin{tabular}{cccccc}
\toprule
Environments & Users & Gestures & Locations & Orientations & Samples \\
\midrule
1st (Classroom) & 9 & 
\makecell[c]{1: Push-Pull; 2: Sweep; 3: Clap; 4: Slide;\\ 5: Draw-O(Horizontal); 6: Draw-Zigzag(Horizontal);\\ 7: Draw-N(Horizontal); 8: Draw-Triangle(Horizontal);\\ 9: Draw-Rectangle(Horizontal)} 
& 5 & 5 & 10,125 \\
\midrule
2nd (Hall) & 4 &  
\makecell[c]{1: Push-Pull; 2: Sweep; 3: Clap; 4: Slide;\\ 5: Draw-O(Horizontal); 6: Draw-Zigzag(Horizontal)} 
& 5 & 5 & 3,000 \\
\midrule
3rd (Office) & 4 & 
\makecell[c]{1: Push-Pull; 2: Sweep; 3: Clap; 4: Slide;\\ 5: Draw-O(Horizontal); 6: Draw-Zigzag(Horizontal)} 
& 5 & 5 & 3,000 \\
\bottomrule
\end{tabular}
\caption{Detailed gesture descriptions of Widar3.0 Dataset}
\label{tab:widar-dataset}
\end{table*}

\begin{table*}[h]
\centering
\begin{tabular}{cccccc}
\toprule
Environments & 
Users & 
Gestures & 
Locations & 
Orientations & 
Samples \\
\midrule
Scene 1 & 30 & 
\makecell[c]{1: Drawing circle; 2: Drawing cross;\\ 
3: Pushing; 4: Pulling; 5: Swiping left;\\ 
6: Swiping right; 7: Swiping up; 8: Swiping down} & 
Changeable & 
Changeable & 
33,000 \\
\cmidrule(r){1-6}
Scene 2 & 3 & 
\makecell[c]{1: Drawing circle; 2: Drawing cross;\\ 
3: Pushing; 4: Pulling; 5: Swiping left;\\ 
6: Swiping right; 7: Swiping up; 8: Swiping down} & 
Changeable & 
Changeable & 
3,300 \\
\cmidrule(r){1-6}
Scene 3 & 3 & 
\makecell[c]{1: Drawing circle; 2: Drawing cross;\\ 
3: Pushing; 4: Pulling; 5: Swiping left;\\ 
6: Swiping right; 7: Swiping up; 8: Swiping down} & 
Changeable & 
Changeable & 
3,300 \\
\cmidrule(r){1-6}
Scene 4 & 3 & 
\makecell[c]{1: Drawing circle; 2: Drawing cross;\\ 
3: Pushing; 4: Pulling; 5: Swiping left;\\ 
6: Swiping right; 7: Swiping up; 8: Swiping down} & 
Changeable & 
Changeable & 
3,300 \\
\bottomrule
\end{tabular}
\caption{Detailed gesture descriptions of XRF55 Dataset}
\label{tab:xrf55-dataset}
\end{table*}

\begin{table*}[h]
\centering
\begin{tabular}{|l|l|l|}
\hline
1. Shaking hands & 20. Playing Ukulele & 39. Standing up \\
2. Hugging & 21. Playing drum & 40. Sitting down \\
3. Handing something to someone & 22. Jumping & 41. Blowing dry hair \\
4. Hitting someone with something & 23. Running & 42. Cutting something \\
5. Choking someone's neck & 24. Turning & 43. Drinking \\
6. Pushing someone & 25. Walking & 44. Putting something on the table \\
7. Kicking someone & 26. Patting on the shoulder & 45. Eating \\
8. Mopping the floor & 27. Foot stamping & 46. Carrying weight \\
9. Combing hair & 28. Shaking head & 47. Brushing teeth \\
10. Weightlifting & 29. Nodding & 48. Using a phone \\
11. Throwing something & 30. Drawing a circle & 49. Hula hooping \\
12. Picking something & 31. Drawing a cross & 50. Jumping rope \\
13. Body weight squats & 32. Pushing & 51. Falling on the floor \\
14. Waving & 33. Pulling & 52. Putting on clothing \\
15. Boxing & 34. Swiping left & 53. Tai Chi \\
16. Jumping jack & 35. Swiping right & 54. Stretching \\
17. Clapping hands & 36. Swiping up & 55. Smoking \\
18. High leg lifting & 37. Swiping down & \\
19. Playing Er-Hu & 38. Wearing a hat & \\
\hline
\end{tabular}
\caption{Detailed behavior descriptions of XRF55 Dataset}
\label{tab:XRF55}
\end{table*}

\end{document}